%% file: mainv1.tex
\pdfoutput=1

\documentclass[11pt]{article}

\usepackage[final]{acl}

\usepackage{times}
\usepackage{latexsym}

\usepackage[T1]{fontenc}

\usepackage[utf8]{inputenc}

\usepackage{microtype}

\usepackage{inconsolata}

\usepackage{graphicx}
\usepackage{array}
\usepackage{multirow}
\usepackage{booktabs}
\usepackage{colortbl}
\usepackage{xcolor}
\usepackage{vcell}
\usepackage{arydshln}
\usepackage{amsmath}
\usepackage{amsfonts}
\usepackage{tabularx}
\usepackage{makecell}
\usepackage[normalem]{ulem}
\usepackage{CJKutf8}
\usepackage{pifont}
\usepackage{arydshln}
\usepackage{tikz}
\usepackage{pgfplots}
\usepackage{inconsolata}

\usepackage{todonotes}
\usepackage[capitalise,noabbrev]{cleveref}
\usepackage{listings}
\usepackage{fancyvrb} 
\makeatletter
\def\adl@drawiv#1#2#3{%
        \hskip.5\tabcolsep
        \xleaders#3{#2.5\@tempdimb #1{1}#2.5\@tempdimb}%
                #2\z@ plus1fil minus1fil\relax
        \hskip.5\tabcolsep}

\definecolor{verbgray}{gray}{0.9}

\usepackage{colortbl}
\definecolor{lightgray}{rgb}{0.7,0.7,0.7}
\def\grayrow{\rowcolor{lightgray}}
\newcommand{\cmark}{{\textbf{\textcolor[rgb]{0.1, 0.5, 0.1}{\ding{51}}}}}
\newcommand{\xmark}{{\textbf{\color{red}{\ding{55}}}}}
\newcommand{\cdashlinelr}[1]{%
  \noalign{\vskip 2pt}   
  \cdashline{#1}[.4pt/2pt] 
  \noalign{\vskip 2pt}   
}
\newcommand{\chinese}[1]{{\begin{CJK*}{UTF8}{gkai} #1 \end{CJK*}}}
\definecolor{light-orange}{HTML}{fee9d4}
\definecolor{light-green}{HTML}{d8f0d3}
\definecolor{light-blue}{HTML}{dae8f5}
\definecolor{light-red}{HTML}{FBC7C4}
\definecolor{set10-red}{HTML}{e41a1c}
\definecolor{set10-blue}{HTML}{377eb8}
\definecolor{set10-green}{HTML}{4daf4a}
\definecolor{bblue}{HTML}{4F81BD}
\definecolor{rred}{HTML}{c4260b}
\definecolor{ggreen}{HTML}{098c1f}
\definecolor{ppurple}{HTML}{9F4C7C}
\definecolor{oorange}{HTML}{F79646}

\DeclareRobustCommand{\hlred}[1]{{\textcolor{rred}{#1}}}
\DeclareRobustCommand{\hlblue}[1]{{\textcolor{bblue}{#1}}}

\usepackage{enumitem}
\setlist[itemize,enumerate]{leftmargin=*}
\usepackage[normalem]{ulem}
\usepackage{pgfplots}
\usetikzlibrary{pgfplots.groupplots}
\pgfplotsset{compat=1.3}
\usepackage{tikz}
\usetikzlibrary{patterns}
\usepackage{xcolor}
\usepackage{todonotes}
\usepackage{listings}
\usepackage{multirow}
\usepackage{graphicx}
\definecolor{CustomBlue}{RGB}{57,83,191}
\usepackage[most]{tcolorbox}

\newtcbox{\clustertab}[1]{on line, box align=base, colback={#1},colframe={#1},size=fbox,arc=2pt,top=-1.5pt, bottom=-1.5pt, left=-1.5pt, right=-1.5pt, boxrule=0pt, enlarge left by=1pt}

\title{M-MAD: Multidimensional Multi-Agent Debate for Advanced \\ 
Machine Translation Evaluation}

\makeatletter
\def\thanks#1{\protected@xdef\@thanks{\@thanks
        \protect\footnotetext{#1}}}
\makeatother

\author{
    Zhaopeng Feng$^{1\heartsuit}$\quad
    Jiayuan Su$^{1\heartsuit}$ \quad
    Jiamei Zheng$^{1}$ \quad
    Jiahan Ren$^{1}$ \quad \\
    \bf Yan Zhang$^{2}$ \quad 
    \bf Jian Wu$^{1}$ \quad
    \bf Hongwei Wang$^{1\dag}$ \quad
    \bf Zuozhu Liu$^{1\dag}$ \\
    $^{1}$Zhejiang University \quad
    $^{2}$National University of Singapore \quad \\
    \texttt{\{zhaopeng.23,jiayuan.23,hongweiwang,zuozhuliu\}@intl.zju.edu.cn} \\
}

\thanks{$^{\heartsuit}$ Equal contribution.} 
\thanks{$^{\dag}$ \space Corresponding author.}

\begin{document}
\maketitle

\begin{abstract}

Recent advancements in large language models (LLMs) have given rise to the LLM-as-a-judge paradigm, showcasing their potential to deliver human-like judgments. However, in the field of machine translation (MT) evaluation, current LLM-as-a-judge methods fall short of learned automatic metrics. In this paper, we propose \textbf{M}ultidimensional \textbf{M}ulti-\textbf{A}gent \textbf{D}ebate (\textbf{M-MAD}), a systematic LLM-based multi-agent framework for advanced LLM-as-a-judge MT evaluation. Our findings demonstrate that M-MAD achieves significant advancements by (1) decoupling heuristic MQM criteria into distinct evaluation dimensions for fine-grained assessments; (2) employing multi-agent debates to harness the collaborative reasoning capabilities of LLMs; (3) synthesizing dimension-specific results into a final evaluation judgment to ensure robust and reliable outcomes. Comprehensive experiments show that M-MAD not only outperforms all existing LLM-as-a-judge methods but also competes with state-of-the-art reference-based automatic metrics, even when powered by a suboptimal model like GPT-4o mini. Detailed ablations and analysis highlight the superiority of our framework design, offering a fresh perspective for LLM-as-a-judge paradigm. Our code and data are publicly available at~\href{https://github.com/SU-JIAYUAN/M-MAD}{https://github.com/SU-JIAYUAN/M-MAD}.
\end{abstract}


\input{table/intro}

\begin{figure*}[ht]
    \centering
    \includegraphics[scale=0.75]{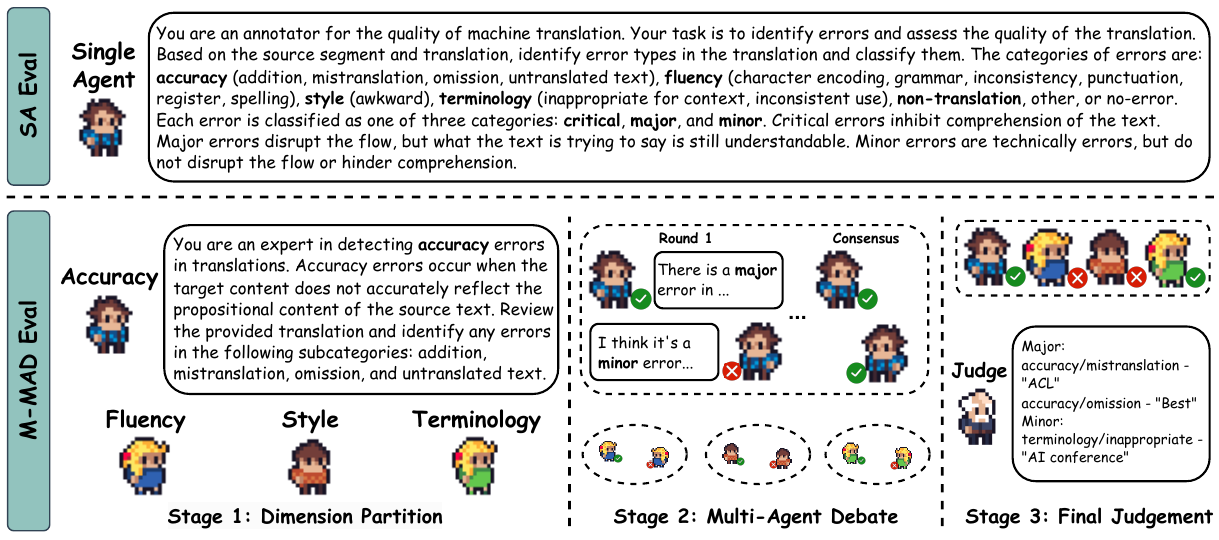}
    \caption{Comparison between Single-Agent (SA) MT Evaluation and Multidimensional Multi-Agent Debate (M-MAD). \textbf{SA:} A single agent evaluates translations using a coupled MQM template (e.g., GEMBA~\citep{kocmi2023gemba}). 
    \textbf{M-MAD:} a) decouple MQM paradigm into four evaluation dimensions; b) multi-agent groups debate within each dimension; c) a judge synthesizes viewpoints from debate groups into the final evaluation. We view M-MAD as analogous to the structure of "a neural network in natural language form". Each stage functions like a layer, each agent serves as a neuron, and their interactions act as hidden states. } 
    \vspace{-3mm}
    \label{fig:method}
\end{figure*}

\section{Introduction}
Evaluating natural language generation systems has long been a challenging task. As the quality of these systems continues to improve, the need for robust and precise evaluation methods has become even more critical~\citep{chang2024survey}. In machine translation (MT) evaluation, learned automatic metrics such as MetricX~\citep{juraska-etal-2023-metricx} and XCOMET~\citep{guerreiro2024xcomet} have achieved state-of-the-art performance in benchmarks like the WMT Metrics shared task~\citep{freitag-etal-2023-results}. However, these model-based metrics require extensive human-annotated datasets for training and rely on reference translations for better accuracy.

The emerging LLM-as-a-judge paradigm offers a promising alternative by leveraging large language models (LLMs) to directly assess or critique generated content~\citep{chiang2023can, zhang-etal-2023-llmaaa, lee2023rlaif, wu2024meta, li2024generation, pavlovic2024effectiveness}.  Early explorations in MT evaluation have shown potential~\citep{kocmi-federmann-2023-large, fernandes-etal-2023-devil, xu2023instructscore, leiter2024prexme}. For example, GEMBA-MQM~\citep{kocmi2023gemba} and EAPrompt~\citep{lu-etal-2024-error} use coupled heuristic Multidimensional Quality Metrics (MQM)~\cite{freitag-etal-2021-experts} prompt to guide LLMs in identifying potential errors, which are aggregated into final quality scores via a weighted scoring system~\citep{freitag-etal-2021-experts}. Despite their comparable performance to learned automatic metrics at the system-level, these methods face significant limitations: (1) they exhibit weaker segment-level performance, lagging behind state-of-the-art model-based automatic metrics (see in Table~\ref{tab:intro}); (2) they rely on coupled heuristic MQM templates, which may introduce biases toward specific error types and further influence the quality score; (3) their single-agent single-step evaluation processes fail to fully exploit the reasoning and collaborative capabilities inherent in LLMs.

In human evaluation, addressing instability and bias often involves dividing tasks into distinct dimensions and fostering collaboration among multiple annotators~\citep{van-der-lee-etal-2019-best,karpinska-etal-2021-perils}. Similarly, in LLM-based multi-agent systems, multi-agent debate has proven effective in generating truthful and factual judgments~\citep{chan2023chateval, khandebating, liang-etal-2024-encouraging, duimproving}. These findings highlight that both human and LLM-based collaborative evaluation processes can achieve greater reliability compared to individual assessments.

Motivated by these observations and insights, we propose the \textbf{M}ultidimensional \textbf{M}ulti-\textbf{A}gent \textbf{D}ebating (\textbf{M-MAD}) framework for MT evaluation. M-MAD operates in three stages: (1) decomposing the heuristic MQM annotation guideline into distinct dimensions for independent LLM-as-a-judge assessments; (2) conducting multi-agent debates within each dimension, harnessing LLMs' inherent knowledge, reasoning, and collaborative abilities; (3) synthesizing the debated outcomes through a final judge agent to produce a comprehensive evaluation judgment.  Comprehensive experiments demonstrate that M-MAD not only maintains the strong system-level performance but also significantly enhances segment-level performance compared to existing LLM-as-a-judge approaches and most automatic metrics. Moreover, even with a suboptimal LLM like GPT-4o mini, M-MAD achieves performance comparable to state-of-the-art reference-based automatic metrics.

Our contributions are summarized as follows:

\begin{itemize}

    \item We introduce the first LLM-based systematic \textbf{M}ultidimensional \textbf{M}ulti-\textbf{A}gent \textbf{D}ebating (\textbf{M-MAD}) framework for MT evaluation. By decoupling the MQM paradigm into distinct evaluation dimensions and integrating multi-agent debates for each dimension, M-MAD enhances the robustness, precision, and reliability of LLM-as-a-judge approaches.
    
    \item M-MAD achieves significant improvements in segment-level performance, surpassing existing LLM-as-a-judge methods while maintaining top-tier system-level accuracy. Its alignment with human judgments outperforms most advanced model-based automatic metrics. Notably, even when powered by a suboptimal LLM (GPT-4o mini), M-MAD delivers results comparable to state-of-the-art reference-based and reference-free metrics, demonstrating the transformative potential of LLM-as-a-judge approaches.

    \item Beyond introducing M-MAD, we conduct comprehensive benchmarking against other multi-agent collaboration frameworks, highlighting the benefits of M-MAD’s dimension-decoupling design. Our thorough ablation studies and analysis explore key components of the framework, such as debating modalities and strategies, emphasizing the importance of fine-grained instructions and error-severity debates in MQM evaluation scenarios. These insights provide a foundation for future research in LLM-based MT evaluation.

\end{itemize}

\section{Multidimensional Multi-Agent Debate Framework}

In this section, we first introduce the background of the MQM paradigm~\citep{freitag-etal-2021-experts}. We then elaborate on the principal stages in M-MAD, including \textit{dimension partition}, \textit{multi-agent debate}, and \textit{final judgment}, as shown in Figure~\ref{fig:method}.

\subsection{Preliminary: Multidimensional Quality Metrics Evaluation Method}
The MQM framework is a flexible human evaluation method designed to assess and categorize translation errors~\citep{burchardt-2013-multidimensional, freitag-etal-2021-experts}. Unlike traditional Direct Assessment (DA), which relies on annotators assigning scores on a 0–100 scale, MQM focuses on identifying specific errors and classifying them based on severity and category. Annotators evaluate translations segment by segment, taking into account the context of the entire document. Each error is assigned a severity level (major or minor) and categorized by type. Spans without errors are marked with neutral severity and no category. This approach provides a more granular and structured evaluation compared to DA. The MQM framework derives quality scores automatically by applying a weighted scheme to the identified errors, considering both their severity and category. Segment-level scores range from 0 (perfect translation) to 25 (worst possible translation), with the overall score calculated as the average across all annotators. For some applications, such as correlating with learned metrics, the scores are negated for consistency. MQM has been shown to align more closely with human judgments than DA, offering a robust and interpretable framework for evaluating MT quality~\citep{freitag-etal-2022-results, zhao2024handcrafted}. More details about MQM are provided in Appendix~\ref{app:mqm}.

\input{table/main_results}

\subsection{Multidimensional Multi-Agent Debate}
In the M-MAD framework, given a translation pair $(x,y)$, there are $d$ evaluation dimensions $\mathcal{D} = \left \{ D_{i}  \right \} _{i=1}^{d} $, where each dimension corresponds to an error category. For each dimension, $n$ agents $\mathcal{A} = \left \{ A_{i}  \right \} _{i=1}^{n} $, each powered by an LLM $L_{i}$, engaged in debating. The debating process lasts for a maximum of $\mathcal{R}$ rounds. During each round $r$, an agent $ A_{i} $ generates a response $\left (s_{i},r\right ) \sim  A_{i}\left ( s_{i} \mid H, P \right ) $, where $H$ is the history of previous messages visible to the agent, and $P$ is the prompt. At the end of the debate, the debating group provides their final viewpoint for its dimension, denoted as $V(D_{i})$. 
After evaluating all dimensions, the final judge agent $J$  gathers all viewpoints $\mathcal{V}=\left \{V(D_{i})\right \}_{i=1}^{d}$ and synthesizes them into an overall evaluation $O(x,y)$.

\noindent{{\textbf{Stage 1: Dimension Partition.}}}
We first decouple the pre-defined MQM guideline into four distinct evaluation dimensions $(d = 4)$: \textit{Accuracy}, \textit{Fluency}, \textit{Style} and \textit{Terminology}. We exclude the rare and easily identifiable error type \textit{non-translation} as well as \textit{locale convention}, which is not related to translation errors, to ensure a fair comparison with GEMBA-MQM~\citep{kocmi2023gemba}. For each dimension, an agent $A_{0}$ performs the initial evaluation $s_{0} \sim A_{0}\left ( s_{0} \mid  P_{0} \right ) $, where $P_{0}$ is the evaluation template designed for this dimension to identify potential error spans, classify them into subcategories, and assign severity.

\noindent{{\textbf{Stage 2: Multi-Agent Debate.}}}
In this stage, we employ "Pro-Con Debate," a formal discussion format where participants present arguments for or against a topic~\citep{johnson1985classroom}. For each evaluation dimension, we assign a two-agent debating group $(n=2)$. Starting with the initial evaluation $s_{0}$, if an error is detected, agent $A_{1}$ supports $s_{0}$, while agent $A_{2}$ holds the opposite. In the first round,  $A_{1}$  generates a statement $(s_{1},1)\sim  A_{1}\left ( s_{1} \mid H, P \right )$, where $H$ includes the initial standpoints of both sides, and $P$ is the debating prompt. $A_{1}$ may provide explanations, reinforce its position, or switch sides. Agent $A_{2}$ then follows the same process to generate $(s_{2},1)$. These statements, $(s_{1},1)$ and $(s_{2},1)$, are added to the history $H$ as context for subsequent rounds. After each round, a consensus checker determines if the agents have reached an agreement. If consensus is achieved, the resulting statement $s_{c}$ becomes the viewpoint $V (Di)$ for this dimension, and the debate ends. If consensus is not reached after  $R$ rounds, the supportive side of $s_{0}$ is taken as $V (Di)$. We conduct extensive experiments to evaluate different debating strategies and their impact in Section~\ref{sec:debate}.


\noindent{{\textbf{Stage 3: Final Judgement.}}}
This stage consists of two steps: viewpoint synthesis and quality score calculation. First, the final judge agent $J$ gathers the set of viewpoints $\mathcal{V}=\left \{V(D_{i})\right \}_{i=1}^{d}$ obtained in Stage 2. Agent $J$ evaluates the validity of each viewpoint, removes redundant information, and synthesizes them into an overall evaluation $O(x,y)$. Next, $O(x,y)$ is used to calculate the translation quality score by counting the number of major and minor errors and applying the following MQM score formula:
\begin{equation}
  MQM \text{ score} = - w_{\text{major}} n_{\text{major}} - w_{\text{minor}} n_{\text{minor}}
\end{equation}
where $n_{major}$ and $n_{minor}$ represent the counts of major and minor errors, respectively, while $w_{major}$ and $w_{minor}$ denote their corresponding severity weights. In line with GEMBA-MQM, EAprompt and WMT Metric Shared Task~\citep{freitag-etal-2023-results}, we set $w_{major} = 5$ and $w_{minor} = 1$.

\section{Experiments}

\subsection{Experimental Setup}

\noindent{{\textbf{Datasets.}}}
Our experiments utilize MQM ratings~\citep{burchardt-2013-multidimensional,freitag-etal-2021-experts} from the WMT 2023 Metrics Shared Task~\citep{freitag-etal-2023-results}, encompassing a total of 45 systems and 68,130 segments across four languages: English (EN), German (DE), Chinese (ZH), and Hebrew (HE). Following the WMT 2023 Metrics Shared Task, we tested three language pairs: EN-DE, ZH-EN, and HE-EN. The human MQM scores serve as the ground truth for evaluating the performance of automatic metrics. Detailed dataset statistics are presented in Table \ref{tab:mqm_stat}.

\noindent{{\textbf{Meta Evaluation.}}}  
We adopt the same meta-evaluation metrics as the WMT 23 Metrics Shared Task, which defines a composite evaluation score based on four scenarios, each contributing equally (weight of 0.25) to the final meta score:  

– system-level pairwise accuracy;

– system-level Pearson correlation;

– segment-level Accuracy-t;

– segment-level Pearson correlation,

All results are computed using MTME~\footnote{\url{https://github.com/google-research/mt-metrics-eval}}, a metric evaluation tool recommended by WMT~\citep{freitag-etal-2023-results} to ensure comparability with other evaluation metrics.

\input{table/ablation}

\noindent{{\textbf{Baseline Metrics.}}} 
We compare M-MAD against the following categories of metrics: 

1) \textbf{Reference-free LLM-as-a-judge methods:} GEMBA-MQM~\citep{kocmi2023gemba} and EAPrompt~\citep{lu-etal-2024-error}; 

2) \textbf{Reference-based metrics:} XCOMET~\citep{guerreiro2024xcomet}, MetricX-23~\citep{juraska-etal-2023-metricx}, BLEURT-20~\citep{sellam2020bleurt}, COMET~\citep{rei2020comet}, MaTESe~\citep{perrella-etal-2022-matese}, and BERTScore~\citep{zhang2019bertscore}; 

3) \textbf{Reference-free metrics:} COMETKiwi~\citep{rei2022cometkiwi}, MetricX-23-QE~\citep{juraska-etal-2023-metricx}, and XCOMET-QE~\citep{guerreiro2024xcomet}.  

We also report results using model-free lexical metrics for completeness, including BLEU~\citep{papineni-etal-2002-bleu} and chrF~\citep{popovic-2015-chrf}. 


\noindent{{\textbf{Implementation Details.}}} 
We use GPT-4o mini as the primary model for all LLM-as-a-judge methods~\footnote{\href{https://platform.openai.com/docs/models/gpt-4o-mini}{https://platform.openai.com/docs/models/gpt-4o-mini}} and set the temperature to 0 to ensure reproducibility. The model's accessibility and API-based interaction streamline its integration into our framework, facilitating both current experiments and future exploration. In M-MAD Stage 1, we adopt a 4-shot demonstration strategy, selecting examples from the WMT 22 MQM dataset. For LLM-as-a-judge methods such as GEMBA-MQM and EAPrompt, we use the templates and examples provided in their respective papers and conduct experiments using the same model and settings as M-MAD. For other baseline automatic metrics, we report their performance using MTME for consistency. More details are in Appendix~\ref{app:prompts}.

\subsection{Main Results}
\textbf{Improvements over LLM-as-a-judge Methods.}
Table~\ref{tab:main_results} demonstrates that M-MAD significantly outperforms existing LLM-as-a-judge frameworks for MT evaluation. On the average Meta score across ZH-EN and EN-DE, M-MAD achieves a 3.8$\%$ improvement over GEMBA-MQM and a 5.4$\%$ improvement over EAPrompt. While maintaining the consistently strong system-level performance, M-MAD achieves notable improvements at the segment level. For instance, on EN-DE segment-level evaluation, M-MAD surpasses GEMBA-MQM by 9.5$\%$ and EAPrompt by 14.4$\%$. 

\noindent{\textbf{Comparison with SoTA Automatic Metrics.}}
We also compare M-MAD with state-of-the-art automatic metrics models, as shown in Table~\ref{tab:main_results}. In the reference-based setting, M-MAD surpasses COMETKiwi by 2.6$\%$ and MetricX-23-QE by 0.9$\%$ in terms of Meta score. Notably, when compared to reference-based metrics, the reference-free and training-free M-MAD outperforms most reference-based metrics, trailing only the XCOMET-Ensemble (XCOMET-XL + XCOMET-XXL). These results underscore the significant potential of M-MAD in advancing the LLM-as-a-judge paradigm for MT evaluation.

\input{table/error_span}

\section{Ablation and Analysis}
We conduct multiple ablations to demonstrate the impact of each component of M-MAD and perform a detailed analysis of our approach.

\subsection{Decoupled Multidimensional Design Brings Significant Improvements}
As shown in Table~\ref{tab:ablation}, Stage 1 (Dimension Partition) contributes the most significant improvements in M-MAD framework. When we remove it from M-MAD, the overall performance dropped by 5.1$\%$. This aligns with the observation of GEMBA-MQM and reinforces our earlier assertion that previous LLM-as-a-judge methods, which rely on coupled multidimensional templates, are suboptimal. Decoupling multidimensional evaluation into separate dimension agents maximizes the potential of each dimension. Dimension partition also provides a more centralized debating focus for Stage 2.

\begin{figure}[ht]
\centering
\input{fig/distribution}
\caption{MQM score distribution for WMT 23 Metrics Shared Task ZH-EN set. HQ: High-quality translations with MQM score $=$ 0. MQ: Medium-quality translations with -5 $<$ MQM score $<$ 0. LQ: Low-quality translations with MQM score $\leq
$ -5.}
\label{fig:score_percentage}
\end{figure}

\subsection{Segment-Level Improvements from More Accurate Error Span Predictions}
We evaluate M-MAD's performance on error span prediction in the WMT 23 ZH-EN Metrics Shared Task using precision, recall, and F1 scores, as shown in Table~\ref{tab:span_score}. Compared to prior LLM-as-a-judge methods, M-MAD demonstrates significant improvements across all metrics, indicating more accurate and reliable error span predictions. As illustrated in Figure~\ref{fig:score_percentage}, both EAPrompt and GEMBA-MQM exhibit a tendency to overestimate error severity compared to the gold human annotations. In contrast, the MQM score distribution generated by M-MAD closely aligns with the gold annotations, outperforming other LLM-as-a-judge approaches. This observation supports our hypothesis that coupled multidimensional instructions can introduce biases toward specific error types, thereby impacting the overall quality score. M-MAD's ability to mitigate these biases contributes to its enhanced segment-level performance, further affirming its effectiveness in MT evaluation.

\input{table/debate_outside}

\subsection{More Precise Debating Approaches Yields Better Evaluation Results}
\label{sec:debate}
Before implementing dimension partitioning, we adapted existing off-the-shelf debating frameworks for MT evaluation based on the coupled evaluation results from GEMBA-MQM. However, as shown in Table~\ref{tab:debating_frame}, directly debating coupled multidimensional results leads to a significant performance drop. We hypothesize that this occurs because agents struggle to identify specific focal points for the debate when addressing multiple dimensions simultaneously. To isolate the influence of multidimensional coupling, we conduct a controlled experiment after Stage 1. In this setup, debaters are allowed to freely debate within distinct dimensions (referred to as "Entirety" in Table~\ref{tab:debate_topic}). Interestingly, the results still show a substantial decline in performance, supporting our hypothesis: unfocused or unstructured debates do not yield positive gains. To identify the most effective debating strategy, we design and test several structured approaches following dimension partitioning, including:

\textbf{Consensus:} Two debaters argue over the severity of errors, with one supporting the initial evaluation and the other opposes it. The debate ends as soon as consensus is reached. If no agreement is reached, the original evaluation is retained.

\textbf{Deliberation:} Debaters present opposing arguments over multiple rounds until the maximum round limit is reached. A judge reviews the transcript and makes the final decision based on the presented arguments.
\input{table/cat_coarse}

\textbf{Interactive Review:} After each round, a reviewer introduces questions to the debaters, who must rebut each other’s arguments and respond to the questions. A judge evaluates the complete transcript and determines the final evaluation.
\begin{figure*}[t]
    \centering
    \includegraphics[scale=0.62]{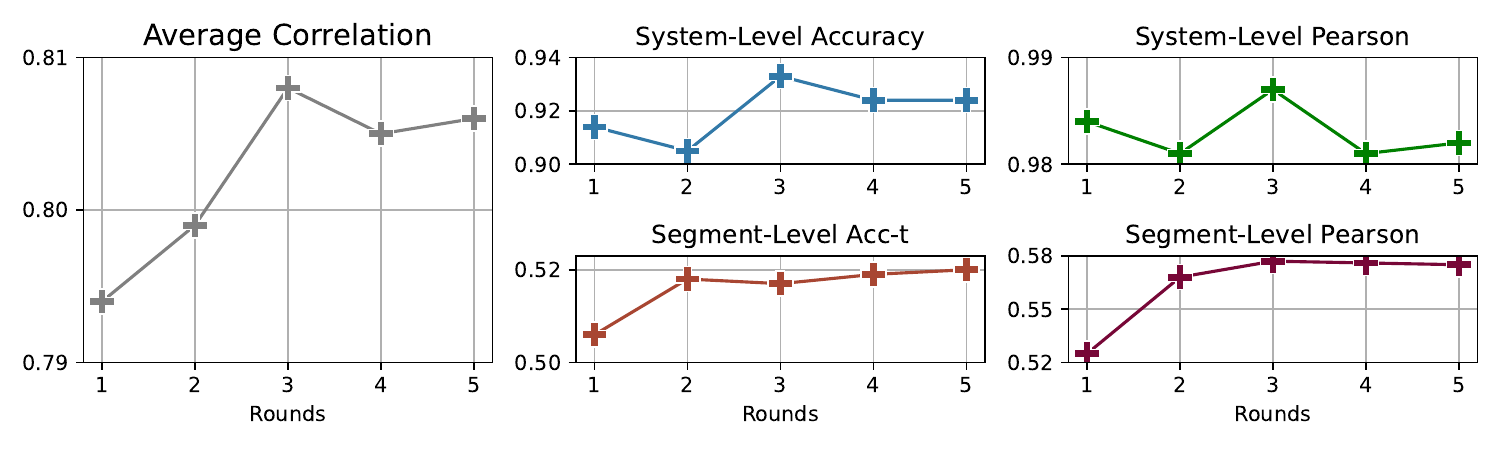}
    \caption{Performance with increased multi-agent debating rounds on WMT 23 ZH-EN.} 
    \vspace{-3mm}
    \label{fig:rounds}
\end{figure*}

\textbf{Consultancy Review:} A debater defends the initial evaluation, engaging directly with a reviewer over multiple rounds. The reviewer collaborates with a judge to determine the final outcome.

\input{table/debate_strategies}

Our experimental results in Table~\ref{tab:debate_strategy} demonstrate that debates conducted within a specific, well-defined dimension yield significantly more stable and reliable outcomes compared to coupled multidimensional debates. Furthermore, the \textit{Consensus} strategy outperforms review-based questioning approaches, highlighting its effectiveness in refining evaluation results. We also investigate the impact of debate focus—comparing error category versus error severity. As shown in Table~\ref{tab:debate_topic}, debates centered on error severity result in greater improvements in meta scores. This outcome is intuitive, as error severity directly influences the final MQM score, thereby playing a critical role in evaluation accuracy. We believe that exploring dynamic scoring mechanisms tailored to different error types could further enhance the effectiveness of MT evaluation in future research.

\subsection{Performance Convergence as Debating Rounds Progress}
To investigate the impact of debating rounds on M-MAD's performance, we conduct experiments with up to five rounds of multi-agent debate. As shown in Figure~\ref{fig:rounds}, system-level performance fluctuates across rounds but consistently peaks at round 3. Segment-level performance, in contrast, exhibits steady improvement during the first three rounds before plateauing. These findings suggest that three rounds of debate strike a balance between refining evaluations and maintaining stability.

\subsection{Case Study}

We provide case studies in Table~\ref{tab:good_case_study} and Table~\ref{tab:bad_case_study} to illustrate M-MAD's effectiveness and limitations. In Table~\ref{tab:good_case_study}, the gold evaluation marks all error spans as minor, whereas EAPrompt and GEMBA-MQM incorrectly identify two major errors, significantly distorting the final MQM scores. Similarly, M-MAD in Stage 1 tends to overestimate errors, with overlapping annotations. These issues are addressed in subsequent stages: the multi-agent debate mechanism in Stage 2 resolves overestimations, while the final judgment in Stage 3 mitigates overlapping annotations. We conceptualize M-MAD as analogous to \textit{"a neural network in natural language form"}. Each stage functions as a layer, agents serve as neurons, and their interactions as hidden states. The flow of language through this internal workflow enables a deeper exploration of the evaluation process. This layered design differentiates M-MAD from traditional LLM-as-a-judge methods and underpins its success.

Nonetheless, as Table~\ref{tab:bad_case_study} demonstrates, M-MAD occasionally diverges from gold annotations. By analyzing the cases with the largest discrepancies, we find instances of potentially mislabeled gold annotations, complicating evaluation. This aligns with the observations of \citet{agrawal-etal-2024-automatic-metrics}, which highlight the increasing difficulty of assessing high-quality translations with automatic metrics as MT systems improve. Developing more robust annotation paradigms beyond MQM could be critical for next-generation MT evaluation frameworks.

\input{table/good_case}

\section{Related Work}
The success of learned MT metrics~\citep{rei2020comet,sellam2020bleurt,rei2022cometkiwi,perrella-etal-2022-matese,naskar-etal-2023-quality,gowda-etal-2023-cometoid,juraska-etal-2023-metricx,guerreiro2024xcomet}, which fine-tune neural network models pretrained on large amounts of data, highlights the importance of leveraging transfer learning to achieve metrics with higher correlation to human judgments. With the advent of LLMs, several studies have explored using LLMs as evaluation metrics. For instance, \citet{dreano-etal-2023-embed} used pre-trained LLaMA2 embeddings to map sentences into vector spaces, employing cosine distance to estimate similarity or dissimilarity. Similarly, \citet{xu2023instructscore} and \citet{fernandes-etal-2023-devil} fine-tuned LLMs such as PaLM~\citep{anil2023palm} and LLaMA~\citep{touvron2023llama} on human-annotated MQM datasets, achieving competitive results for specific language pairs. Other research focuses on directly leveraging the internal knowledge of LLMs. Key comparisons in our study, such as EAPrompt~\citep{lu-etal-2024-error} and GEMBA~\citep{kocmi-federmann-2023-large,kocmi2023gemba}, showcase different methodologies for LLM-based evaluation. EAPrompt employs error-severity-focused prompts to guide ChatGPT in the evaluation process, while GEMBA adopts heuristic templates inspired by DA and MQM guidelines, mimicking human annotators to generate evaluation results. Beyond these core comparisons, \citet{leiter2024prexme} explored large-scale prompt engineering to improve evaluation quality. However, challenges persist. \citet{agrawal-etal-2024-automatic-metrics} highlighted that current automatic metrics often fail in high-quality translation scenarios, where distinguishing subtle differences becomes increasingly difficult. Additionally, \citet{deutsch-etal-2023-ties} revealed systemic inconsistencies in the meta-evaluation of metrics, with noticeable discrepancies between segment-level and system-level performance. Our work presents a comprehensive multi-agent collaboration framework for LLM-as-a-judge approaches, offering a novel perspective to advance MT evaluation methodologies.


\section{Conclusion}

In this work, we introduce M-MAD, an advanced multi-agent collaboration framework that elevates LLM-as-a-judge methods for MT evaluation. M-MAD employs a three-stage workflow that decouples MQM criteria into distinct dimensions, conducts multi-agent debates within each dimension, and synthesizes the outcomes into a final robust evaluation. Experimental results demonstrate that M-MAD not only outperforms existing LLM-as-a-judge methods but is also competitive with the advanced reference-free and reference-based automatic metrics. Detailed ablations and analyses reveal the mechanisms behind M-MAD’s success, highlight its ability to leverage fine-grained evaluations and collaborative reasoning to enhance reliability and precision. We believe M-MAD offers a groundbreaking approach for advancing MT evaluation and sets the stage for developing next-generation LLM-as-a-judge frameworks.


\section*{Limitations}
Due to the meta-evaluation's unique complexity, the token consumption required by M-MAD is substantial. As a result, we were unable to afford cutting-edge models like GPT-4o, o1, or Claude-3.5 Sonnet for our experiments. This constraint limits our ability to explore the upper performance bounds of the framework with these advanced models. Our current research primarily focuses on homogeneous groups of LLMs, which might not capture the full potential of multi-agent collaboration. Future research could explore heterogeneous groups of LLMs, where stronger and weaker, close source and open source models cooperate, offering interesting insights into how various models with different strengths and weaknesses can complement each other. 

\bibliography{custom}

\clearpage

\appendix
\section{Datasets}
As shown in Table~\ref{tab:mqm_stat}, the dataset consists of three language pairs from the WMT23 Metrics Shared Task: English-German (En-De), Hebrew-English (He-En), and Chinese-English (Zh-En). Each language pair is evaluated using a variety of systems, including GPT4-5shot, NLLB\_Greedy, and ONLINE-A, among others, and incorporates multiple reference translations, such as refA, refB, and synthetic\_{ref}. This diverse set of systems and references ensures a thorough evaluation of machine translation models across different language pairs.

\section{Implementation Prompts}
\label{app:prompts}
Our method uses few-shot examples, as demonstrated in Figure~\ref{app:acc_examples} and Figure~\ref{app:fluen_examples}. To reproduce GEMBA-MQM, we use the official prompt and examples provided in their papers, as shown in Figure~\ref{fig:prompt} and Figure~\ref{gemba_examples}. For EAPrompt, we also use the official prompts and present the prompt for DE-EN in Figure~\ref{fig:prompt2}.

\section{Multidimensional Quality Metrics (MQM)}
\label{app:mqm}
The annotator guidelines are provided in Table~\ref{tab:mqm-guidelines}. Possible error categories are displayed in Table~\ref{tab:mqm-hierarchy}. 

\section{Results for WMT 23 HE-EN}
Due to page limitations, we provide the results for WMT 23 He-En in Table~\ref{tab:he_results}. Aligned with the main results of Zh-En and En-De, M-MAD is also superior among all metrics in the He-En task, which shows the effectiveness of M-MAD.

\input{table/appendix_he}
\section{Analysis of Incorrect Gold Annotations}
\input{table/bad_case}
In Table~\ref{tab:bad_case_study}, this case represents one of several instances where gold annotations, labeled by human annotators, contain errors. The gold annotations incorrectly flag several pronouns ("you," "she," "her") as errors, despite their correct usage in the translation, leading to unjustified point deductions. Additionally, the gold annotations misjudge grammatical and fluency issues, marking phrases like "you don't get in touch with her, she gets in touch with you" as awkward or incorrect. These errors are consistent across multiple systems, all annotated by human annotators, indicating that the gold annotations themselves may be unreliable. This case highlights the importance of improving the quality of gold annotations and the risks of relying solely on them for evaluation. It also demonstrates that our system’s annotations offer a more reasonable and accurate assessment.

\input{table/dataset}
\label{app:mqm_description}
\begin{table*}[!htb]\centering
\scalebox{1.00}{
\noindent\fbox{%
\parbox{1.0\textwidth}{%

You will be assessing translations at the segment level, where a segment may contain one or more sentences. Each segment is aligned with a corresponding source segment, and both segments are displayed within their respective documents. Annotate segments in natural order, as if you were reading the document. You may return to revise previous segments.\\

Please identify all errors within each translated segment, up to a maximum of five. If there are more than five errors, identify only the five most severe. If it is not possible to reliably identify distinct errors because the translation is too badly garbled or is unrelated to the source, then mark a single \emph{Non-translation} error that spans the entire segment.\\
 
To identify an error, highlight the relevant span of text, and select a category/sub-category and severity level from the available options. (The span of text may be in the source segment if the error is a source error or an omission.) When identifying errors, please be as fine-grained as possible. For example, if a sentence contains two words that are each mistranslated, two separate mistranslation errors should be recorded. If a single stretch of text contains multiple errors, you only need to indicate the one that is most severe. If all have the same severity, choose the first matching category listed in the error typology (eg, \emph{Accuracy}, then \emph{Fluency}, then \emph{Terminology}, etc).\\
 
Please pay particular attention to document context when annotating. If a translation might be questionable on its own but is fine in the context of the document, it should not be considered erroneous; conversely, if a translation might be acceptable in some context, but not within the current document, it should be marked as wrong.\\
 
There are two special error categories: \emph{Source error} and \emph{Non-translation}. Source errors should be annotated separately, highlighting the relevant span in the source segment. They do not count against the five-error limit for target errors, which should be handled in the usual way, whether or not they resulted from a source error. There can be at most one \emph{Non-translation} error per segment, and it should span the entire segment. No other errors should be identified if \emph{Non-Translation} is selected.}
}}
\caption{MQM annotator guidelines}
\label{tab:mqm-guidelines}
\end{table*}

\newpage
\begin{table*}\centering
\scalebox{0.80}{
\begin{tabular}{ll|l}\toprule
\multicolumn{2}{l|}{Error Category} & Description \\
\midrule
Accuracy & Addition    & Translation includes information not present in the source. \\
    & Omission         & Translation is missing content from the source. \\
    & Mistranslation   & Translation does not accurately represent the source.\\
    & Untranslated text & Source text has been left untranslated. \\
\midrule
Fluency & Punctuation   & Incorrect punctuation (for locale or style). \\
    & Spelling          & Incorrect spelling or capitalization. \\
    & Grammar           & Problems with grammar, other than orthography. \\
    & Register          & Wrong grammatical register (eg, inappropriately informal pronouns). \\
    & Inconsistency     & Internal inconsistency (not related to terminology). \\
    & Character encoding          & Characters are garbled due to incorrect encoding. \\
\midrule
Terminology & Inappropriate for context & Terminology is non-standard or does not fit context.\\
            & Inconsistent use & Terminology is used inconsistently.\\
\midrule
Style & Awkward & Translation has stylistic problems.\\
\midrule
Locale & Address format & Wrong format for addresses.\\
convention  & Currency format & Wrong format for currency.\\
    & Date format & Wrong format for dates. \\
    & Name format & Wrong format for names. \\
    & Telephone format & Wrong format for telephone numbers. \\
    & Time format & Wrong format for time expressions. \\
\midrule
Other & & Any other issues. \\
\midrule
Source error & & An error in the source. \\
\midrule
Non-translation & & Impossible to reliably characterize distinct errors.\\
\bottomrule
\multicolumn{3}{c}{}\\
\end{tabular}
}
\caption{MQM hierarchy.}
\label{tab:mqm-hierarchy}
\end{table*}

\begin{figure*}[htb]
{\footnotesize
    \begin{Verbatim}[commandchars=+\[\]]
    (System) You are an annotator for the quality of machine translation. Your task is to identify 
    errors and assess the quality of the translation.
    
    (user) +textbf[{source_language}] source:\n
    ```+textbf[{source_segment}]```\n
    +textbf[{target_language}] translation:\n
    ```+textbf[{target_segment}]```\n
    \n
    Based on the source segment and machine translation surrounded with triple backticks, identify 
    error types in the translation and classify them. The categories of errors are: accuracy 
    (addition, mistranslation, omission, untranslated text), fluency (character encoding, grammar, 
    inconsistency, punctuation, register, spelling),  
    style (awkward), terminology (inappropriate  for context, inconsistent use), non-translation, 
    other, or no-error.\n
    Each error is classified as one of three categories: critical, major, and minor. 
    Critical errors inhibit comprehension of the text. Major errors disrupt the flow, but what 
    the text is trying to say is still understandable. Minor errors are technically errors, 
    but do not disrupt the flow or hinder comprehension.

    (assistant) +textbf[{observed error classes}]
    \end{Verbatim}
}
\caption{The general prompt for GEMBA-MQM,  with the ``(user)'' and ``(assistant)'' section repeated for each few-shot example.}
\label{fig:prompt}
\end{figure*}

\begin{figure*}[htb]
{\footnotesize
    \begin{Verbatim}
English source: I do apologise about this, we must gain permission from the account holder to discuss
an order with another person, I apologise if this was done previously, however, I would not be able 
to discuss this with yourself without the account holders permission.
German translation: Ich entschuldige mich dafür, wir müssen die Erlaubnis einholen, um eine Bestellung 
mit einer anderen Person zu besprechen. Ich entschuldige mich, falls dies zuvor geschehen wäre, aber 
ohne die Erlaubnis des Kontoinhabers wäre ich nicht in der Lage, dies mit dir involvement.
MQM annotations: 
Critical: 
no-error
Major:
accuracy/mistranslation - "involvement"
accuracy/omission - "the account holder"
Minor:
fluency/grammar - "wäre"
fluency/register - "dir"
    \end{Verbatim}
    
    \begin{Verbatim}
English source: Talks have resumed in Vienna to try to revive the nuclear pact, with both sides 
trying to gauge the prospects of success after the latest exchanges in the stop-start negotiations.
Czech transation: Ve Vídni se ve Vídni obnovily rozhovory o oživení jaderného paktu, přičemže obě 
partaje se snaží posoudit vyhlídky na úspěch po posledních výměnách v jednáních.
MQM annotations: 
Critical:
no-error
Major:
accuracy/addition - "ve Vídni"
accuracy/omission - "the stop-start"
Minor:
terminology/inappropriate for context - "partaje"
    \end{Verbatim}
    
\begin{CJK}{UTF8}{gbsn}
    \begin{Verbatim}
Chinese source: 大众点评乌鲁木齐家居商场频道为您提供高铁居然之家地址，电话，营业时间等最新商户信息，
找装修公司，就上大众点评
English translation: Urumqi Home Furnishing Store Channel provides you with the latest business 
information such as the address, telephone number, business hours, etc., of high-speed rail, and 
find a decoration company, and go to the reviews.
MQM annotations: 
Critical:
accuracy/addition - "of high-speed rail"
Major:
accuracy/mistranslation - "go to the reviews"
Minor:
style/awkward - "etc.,"
    \end{Verbatim}
\end{CJK}
}

\caption{Three examples used for GEMBA-MQM.}
\label{gemba_examples}
\end{figure*}

\begin{figure*}[t]
{\footnotesize
    \begin{Verbatim}[commandchars=+\[\]]
    (user) Source: They were addressed to her son, who has autism and lives in a private care facility,
    she said. But instead of her son's name inside when you opened them, the letters said 
    Dear Maine's Department of Health and Human Services -- in Cincinnati, she told local media.
    
    Translation: Sie wurden an ihren Sohn gerichtet, der Autismus hat und in einer privaten 
    Pflegeeinrichtung lebt, sagte sie. Aber anstelle des Namens ihres Sohnes im Inneren, als 
    Sie sie öffneten, sagten die Briefe Dear Maine 's Department of Health and Human Services -- 
    in Cincinnati, sagte sie den lokalen Medien. 
    
    Based on the given source, identify the major and minor errors in this translation. Note that 
    Major errors refer to actual translation or grammatical errors, and Minor errors refer to smaller
    imperfections, and purely subjective opinions about the translation. 
    Count the number of major and minor errors and compute the final score for this translation. 
    Deduct 5 points for each major error. Deduct 1 point for each minor error. If the translation has 
    no errors, its score will be 0. Remember to output the calculated score within <score></score> 
    tags at the end.
    
    Major errors:
    (1) “Sie” – Mistranslation
    (2) “Dear Maine 's Department of Health and Human Services” – Untranslated text
    
    Minor errors:
    (1) “sagten” – Mistranslation
    (2) “im Inneren” – Mistranslation
    (3) “Briefe ,,” – Omission
    (4) “wurden” – Grammar
    (5) “im Inneren, als Sie sie öffneten, sagten die Briefe” – Awkward Style
    
    Based on the above evaluation, The final score for this translation is -5-5-1-1-1-1-1=-15. 
    <score>-15</score>

    (user) Use the template above to answer the following question:
    Source: {source_segment}
    Translation: {target_segment}
    Based on the given source, identify the major and minor errors in this translation. Note that
    Major errors refer to actual translation or grammatical errors, and Minor errors refer to 
    smaller imperfections, and purely subjective opinions about the translation. 
    Count the number of major and minor errors and compute the final score for this translation.
    Deduct 5 points for each major error. Deduct 1 point for each minor error. If the translation 
    has no errors, its score will be 0. Remember to output the calculated score within <score></score> 
    tags at the end.
    \end{Verbatim}
}
\caption{The prompt for EAPrompt, demonstrated here with an example for EN-DE.}
\label{fig:prompt2}
\end{figure*}

\begin{figure*}[t]
{\footnotesize
    \begin{Verbatim}[commandchars=+\[\]]
    (system) You are an annotator for the quality of machine translation. You will be assessing 
    translations at the segment level, where a segment may contain one or more sentences. Each 
    segment is aligned with a corresponding source segment. Please identify all errors within 
    the specified category within each translated segment. To identify an error, highlight the 
    relevant span of text, and select a category/sub-category and severity level from the available
    options. (The span of text may be in the source segment if the error is a source error or an 
    omission.) When identifying errors, please be as fine-grained as possible. For example, if a
    sentence contains two words that are each mistranslated, two separate mistranslation errors 
    should be recorded. If a single stretch of text contains multiple errors, you only need to 
    indicate the one that is most severe. For severity, Major represents errors that may confuse
    or mislead the reader due to a significant change in meaning or because they appear in a 
    visible or important part of the content; Minor represents errors that don’t lead to loss of
    meaning and wouldn’t confuse or mislead the reader but would be noticed, would decrease 
    stylistic quality, fluency or clarity, or would make the content less appealing. In case 
    when it is not possible to reliably identify distinct errors because the translation is too 
    badly garbled or is unrelated to the source, then mark a special category, called 
    non-translation, that spans the entire segment. There can be at most one non-translation
    error per segment, and it should span the entire segment. No other errors should be
    identified if non-translation is selected.

    (user) You are an accuracy errors detection expert for translations. Please check the 
    translation for the following subcategories of accuracy errors:\n\n1. **Accuracy 
    Addition**: The translation includes information not present in the source.\n2. **Omission
    translation**: The translation is missing content from the source.\n3. **Mistranslation**:
    The translation does not accurately represent the source.\n4. **Untranslated Text**: Source 
    text has been left untranslated.\n\nPlease analyze the following segment pair and annotate 
    errors.\n\n##src_lng## source:\n##source_segment##\n##tgt_lng## translation:\n##target_segment##
    \n\nProvide your annotations in JSON format as follows: {\"annotations\":{\"error_span\":
    (if non-translation error is selected, provide 'all'; otherwise, the error_span must be
    chosen from within the translated segment), \"category\":\"(accuracy/{subcategory} or 
    non-translation)\", \"severity\":\"(minor or major)\", \"is_source_error\":\"(yes or no)\"},...}
    
    (assistant) ... (1st example)
    
    (user) You are an accuracy errors...
    
    (assistant) ... (2nd example)
    
    (user) You are an accuracy errors...
    
    (assistant) .... (3rd example)
    
    (user) You are an accuracy errors....
    
    (assistant) .... (4th example)
    
    (user) You are an accuracy errors....

    \end{Verbatim}
}
\caption{The prompt for M-MAD Stage 1, demonstrated here with an example for Accuracy agent.}
\label{fig:prompt2}
\end{figure*}

\begin{figure*}[t]
{\footnotesize
    \begin{Verbatim}[commandchars=+\[\]]
    **Round 1**
    Debater A:
    
    (system) You are an expert in detecting accuracy errors in translations. Accuracy errors
    occur when the target content does not accurately reflect the propositional content of 
    the source text. Review the provided translation and identify any errors in the following
    subcategories: addition, mistranslation, omission, and untranslated text. If no errors are 
    found, return {\"annotations\": []}. If errors are found, list each error with the following 
    details: the exact error span, the subcategory, and the severity (major or minor). Major 
    errors significantly impact meaning and may confuse or mislead the reader. Minor errors have 
    a slight impact, but do not cause loss of meaning or confusion. In case when it is not possible
    to reliably identify distinct errors because the translation is too badly garbled or is 
    unrelated to the source, then mark a special category, called non-translation. No other errors 
    should be identified if non-translation is selected.

    (user) ##src_lng## source:\n##source_segment##\n##tgt_lng## translation:\n##target_segment##
    
    (assistant) ##annotations##
    
    (user) These are the annotations from the other agent: ##the other agent's response## Given 
    two different answers, think about it again. Examine your annotations and the other agent's 
    annotations step by step. When the severity is hard to decide, please lean toward "minor." 
    Only assign "major" if it significantly impacts the meaning. Avoid assigning 'non-translation'
    unless absolutely necessary. Provide your answer in the following JSON format at the end of your 
    response: ```json\n{\"annotations\":{\"error_span\": {error span in translated segment}, 
    \"category\":\"{category}/{subcategory}\", \"severity\":\"{minor or major}\", \"is_source_error\":
    \"{yes or no}\"},...}\n. If no errors are annotated, use the json format: ```json\n{"annotations": 
    []}\n   
    
    (assistant) (Debater A's response)
    -----------------------------------------------------------------------------------------------
    Debater B:
    
    (system) You are an expert in detecting accuracy errors in translations. Accuracy errors
    occur when the target content does not accurately reflect the propositional content of 
    the source text. Review the provided translation and identify any errors in the following
    subcategories: addition, mistranslation, omission, and untranslated text. If no errors are 
    found, return {\"annotations\": []}. If errors are found, list each error with the following 
    details: the exact error span, the subcategory, and the severity (major or minor). Major 
    errors significantly impact meaning and may confuse or mislead the reader. Minor errors have 
    a slight impact, but do not cause loss of meaning or confusion. In case when it is not possible
    to reliably identify distinct errors because the translation is too badly garbled or is 
    unrelated to the source, then mark a special category, called non-translation. No other errors 
    should be identified if non-translation is selected.

    (user) ##src_lng## source:\n##source_segment##\n##tgt_lng## translation:\n##target_segment##
    
    (assistant) ##annotations## (All "major" errors are re-annotated to "minor" errors.)
    
    (user) These are the annotations from the other agent: ##the other agent's response## Given 
    two different answers, think about it again. Examine your annotations and the other agent's 
    annotations step by step. When the severity is hard to decide, please lean toward "minor." 
    Only assign "major" if it significantly impacts the meaning. Avoid assigning 'non-translation'
    unless absolutely necessary. Provide your answer in the following JSON format at the end of your 
    response: ```json\n{\"annotations\":{\"error_span\": {error span in translated segment}, 
    \"category\":\"{category}/{subcategory}\", \"severity\":\"{minor or major}\", \"is_source_error\":
    \"{yes or no}\"},...}\n. If no errors are annotated, use the json format: ```json\n{"annotations": 
    []}\n   
    
    (assistant) (Debater B's response)
    -----------------------------------------------------------------------------------------------
    Consensus checker:
    
    (user) Compare the error annotations provided by two agents for the same machine-translated segment. 
    Determine if the annotations are essentially consistent. The first agent annotations are: 
    {first_annotations}; the other agent annotations are: {second_annotations}. Return \"yes\" if they 
    are consistent, or \"no\" if they are inconsistent. Provide no additional output.
    
    (assistant) ... (If yes, return Debater A's response; if no, continue to the next round...)

    \end{Verbatim}
}
\caption{The prompt for M-MAD Stage 2, demonstrated here with an example for Accuracy agent.}
\label{fig:prompt2}
\end{figure*}

\begin{figure*}[t]
{\footnotesize
    \begin{Verbatim}[commandchars=+\[\]]
    (system) You are a judge for the translation error annotations, given the translation pair 
    and annotations from previous rounds. Your task is to integrate them and remove repeated 
    annotations if any, where if a single error_span contains multiple errors, indicate only 
    the one that is the most severe, and if some errors have the same severity, choose the 
    first matching category listed in the error typology (accuracy, then fluency, then terminology,
    then style). But please note this rule is only applied when a single error_span contains 
    multiple errors. However, if it is not possible to reliably identify distinct errors because
    the translation is too badly garbled or is unrelated to the source, then mark a special category,
    called non-translation, that spans the entire segment. There can be at most one non-translation
    error per segment, and it should span the entire segment. No other errors should be identified
    if non-translation is selected.

    (user) Regarding the translation pair \n\n##src_lng## source:\n##source_segment##\n##tgt_lng## 
    translation:\n##target_segment##\n\nFrom previous annotations, we have the accuracy errors detection 
    expert annotations: \n\n##accuracy_annotations##; the fluency errors detection expert annotations: 
    \n\n##fluency_annotations##; the terminology errors detection expert annotations: \n\n
    ##term_annotations##; and the style errors detection expert annotations: \n\n##style_annotations##.
    \n\nBased on the above information, output your analyses and the final annotations in JSON format 
    as follows: {\"analysis\":(first, judge if it is non-translation error, if yes, explain responsibly
    why it is; if no, explain how do you use the rule when a single error_span contains multiple errors
    to output final annotations), \"annotations\":{\"error_span\":(if non-translation error is selected,
    provide 'all'; otherwise, the error_span must be chosen from within the translated segment),
    \"category\":\"({category}/{subcategory} or non-translation)\", \"severity\":\"(minor or major)\", 
    \"is_source_error\":\"(yes or no)\"},...}

    \end{Verbatim}
}
\caption{The prompt for M-MAD Stage 3.}
\label{fig:prompt2}
\end{figure*}

\begin{figure*}[t]
{\footnotesize
\begin{CJK}{UTF8}{gbsn}
    \begin{Verbatim}
    Chinese source: 工厂直销生产，欢迎代理批发！
    English translation: Factory direct production, welcome agent wholesale!
    MQM annotations: 
    {"annotations":[{"error_span": "Factory direct production", "category": "accuracy/mistranslation", 
    "severity": "major"}, {"error_span": "agent wholesale", "category": "accuracy/mistranslation", 
    "severity": "major"}]}
        \end{Verbatim}
    \end{CJK}    
    \begin{CJK}{UTF8}{gbsn}
        \begin{Verbatim}
    Chinese source: 性能稳定，四个出风口散热没问题，值得推荐。
    English translation: The performance is stable, and the heat dissipation of the four air outlets is 
    no problem, which is worth recommending.
    MQM annotations: 
    {"annotations": []}
        \end{Verbatim}
    \end{CJK} 
        
    \begin{CJK}{UTF8}{gbsn}
        \begin{Verbatim}
    Chinese source: 小编为大家带来洁面乳霜哪款好？
    English translation: Xiaobian brings cleansing cream to everyone. Which one is good?
    MQM annotations: 
    {"annotations": [{"error_span": "Xiaobian", "category": "accuracy/mistranslation", "severity":
    "minor"}]}
        \end{Verbatim}
    \end{CJK} 
    
    \begin{CJK}{UTF8}{gbsn}
        \begin{Verbatim}
    Chinese source: 赴宴来支雲，留下好心情！
    English translation: I'm Xiao Yun, looking forward to meeting you again!
    MQM annotations: 
    {"annotations": [{"error_span": "all", "category": "non-translation", "severity": "major", 
    "is_source_error": "no"}]}
        \end{Verbatim}
    \end{CJK} 
    }
\caption{Our examples used in Stage 1 for the Accuracy agent in Zh-En few-shot prompting.}
\label{app:acc_examples}
\end{figure*}

\begin{figure*}[htb]
{\footnotesize
\begin{CJK}{UTF8}{gbsn}
    \begin{CJK}{UTF8}{gbsn}
        \begin{Verbatim}
    Chinese source: 这是我除了驾驶证之外一本觉得重要的技能证书之一！
    English translation: This is one of the skills certificates that I think are important besides
    my driver's license!
    MQM annotations: 
    {"annotations": [{"error_span": "are", "category": "fluency/grammar", "severity": "minor"}]}
        \end{Verbatim}
    \end{CJK}

    \begin{CJK}{UTF8}{gbsn}
        \begin{Verbatim}
    Chinese source: 餐厅表示已经煮好食物半个多小时了。
    English translation: The restaurant said it had cooked the food for more than half an hour.
    MQM annotations: 
    {"annotations": []}
        \end{Verbatim}
    \end{CJK} 
        
    \begin{Verbatim}
    Chinese source: 感观不错，色彩还原好，质量可以！
    English translation: Good sense, good color restoration and good quality!
    MQM annotations: 
    {"annotations": [{"error_span": " and", "category": "fluency/punctuation", "severity": "minor"}]}
        \end{Verbatim}
    \end{CJK}   
    
    \begin{CJK}{UTF8}{gbsn}
        \begin{Verbatim}
    Chinese source: 赴宴来支雲，留下好心情！
    English translation: I'm Xiao Yun, looking forward to meeting you again!
    MQM annotations: 
    {"annotations": [{"error_span": "all", "category": "non-translation", "severity": "major", 
    "is_source_error": "no"}]}
        \end{Verbatim}
    \end{CJK} 
    }
\caption{Our examples used in Stage 1 for the Fluency agent in Zh-En few-shot prompting.}
\label{app:fluen_examples}
\end{figure*}

\begin{figure*}[htb]
{\footnotesize
\begin{CJK}{UTF8}{gbsn}
    \begin{CJK}{UTF8}{gbsn}
        \begin{Verbatim}
    Chinese source: 散热效果：散热非常好，基本无热度 轻薄程度：方便携带 外观材质：Thinkpad传统设计，
    满意
    English translation: Heat dissipation effect: very good heat dissipation, basically no heat, 
    light and thin degree: easy to carry appearance material: Thinkpad traditional design, 
    satisfactory
    MQM annotations: 
    {"annotations": []}
        \end{Verbatim}
    \end{CJK}

    \begin{CJK}{UTF8}{gbsn}
        \begin{Verbatim}
    Chinese source: 有习近平总书记作为党中央的核心、全党的核心领航掌舵，有习近平新时代中国特色社会
    主义思想科学指引，有全党全国各族人民团结一心、顽强奋斗，我们就一定能够战胜各种艰难险阻，在全面
    建设社会主义现代化国家新征程上创造新的时代辉煌、铸就新的历史伟业。
    English translation: With General Secretary Xi as the core of the CPC Central Committee and 
    the core of the whole party at the helm, With the scientific guidance of Xi's thought of 
    socialism with Chinese characteristics in the new era, and the unity and tenacious struggle of 
    the whole Party and the people of all ethnic groups, we will be able to overcome all kinds 
    of difficulties and obstacles, create new era glory and create new historical great achievements 
    in the new journey of building a socialist modernized country in an all-round way.
    MQM annotations: 
    {"annotations": [{"error_span": "thought of socialism with Chinese characteristics in the 
    new era", "category": "terminology/inappropriate for context", "severity": "minor"}]}
        \end{Verbatim}
    \end{CJK} 
        
    \begin{Verbatim}
    Chinese source: A字短裙会很显瘦，比较适合胖女孩，上衣要选择紧身一些的，形成层次感会更显瘦。
    English translation: A-line skirt will be very thin, which is more suitable for fat girls. 
    The jacket should be tight, forming a sense of hierarchy will be thinner.
    MQM annotations: 
    {"annotations": []}
        \end{Verbatim}
    \end{CJK}   
    
    \begin{CJK}{UTF8}{gbsn}
        \begin{Verbatim}
    Chinese source: 赴宴来支雲，留下好心情！
    English translation: I'm Xiao Yun, looking forward to meeting you again!
    MQM annotations: 
    {"annotations": [{"error_span": "all", "category": "non-translation", "severity": "major", 
    "is_source_error": "no"}]}
        \end{Verbatim}
    \end{CJK} 
    }
\caption{Our examples used in Stage 1 for the Terminology agent in Zh-En few-shot prompting.}
\label{app:fluen_examples}
\end{figure*}

\begin{figure*}[htb]
{\footnotesize
\begin{CJK}{UTF8}{gbsn}
    \begin{CJK}{UTF8}{gbsn}
        \begin{Verbatim}
    Chinese source: 商家服务态度好！
    English translation: Good service attitude of merchants!
    MQM annotations: 
    {"annotations": [{"error_span": "merchants", "category": "style/awkward", "severity": "minor"}]}
        \end{Verbatim}
    \end{CJK}

    \begin{CJK}{UTF8}{gbsn}
        \begin{Verbatim}
    Chinese source: 打拐执法，首要目标是寻回被拐的孩子，保护其人身安全、自由。
    English translation: The primary goal of law enforcement against abduction is to find abducted
    children and protect their personal safety and freedom.
    MQM annotations: 
    {"annotations": [{"error_span": "merchants", "category": "style/awkward", "severity": "minor"}]}
        \end{Verbatim}
    \end{CJK} 
        
    \begin{Verbatim}
    Chinese source: 从党的百年奋斗史中汲取奋进力量
    English translation: Draw the strength of forging ahead from the party's hundred-year struggle 
    history
    MQM annotations: 
    {"annotations": []}
        \end{Verbatim}
    \end{CJK}   
    
    \begin{CJK}{UTF8}{gbsn}
        \begin{Verbatim}
    Chinese source: 赴宴来支雲，留下好心情！
    English translation: I'm Xiao Yun, looking forward to meeting you again!
    MQM annotations: 
    {"annotations": [{"error_span": "all", "category": "non-translation", "severity": "major", 
    "is_source_error": "no"}]}
        \end{Verbatim}
    \end{CJK} 
    }
\caption{Our examples used in Stage 1 for the Style agent in Zh-En few-shot prompting.}
\label{app:fluen_examples}
\end{figure*}

\begin{figure*}[htb]
{\footnotesize
\begin{CJK}{UTF8}{gbsn}
    \begin{CJK}{UTF8}{gbsn}
        \begin{Verbatim}
    English source: Good Afternoon, thank you for getting in contact with us today, you're through 
    to #NAME#.
    German translation: Guten Tag, vielen Dank, dass Sie sich heute mit uns in Verbindung gesetzt 
    haben, Sie sind durch zu #NAME#.
    MQM annotations: 
    {"annotations":[{"error_span": "Sie sind durch zu", "category": "accuracy/mistranslation", 
    "severity": "major", "is_source_error": "no"}]}
        \end{Verbatim}
    \end{CJK}

    \begin{CJK}{UTF8}{gbsn}
        \begin{Verbatim}
    English source: I'm unable to make any changes once the order has been placed however, when the 
    rider leaves the restaurant you will be able to contact them through the app.
    German translation: Ich kann keine Änderungen vornehmen, sobald die Bestellung aufgegeben wurde, 
    aber wenn der Fahrer das Restaurant verlässt, können Sie ihn über die App kontaktieren.
    MQM annotations: 
    {"annotations": []}
        \end{Verbatim}
    \end{CJK} 
        
    \begin{Verbatim}
    English source: It is a rear occasion that it happens.
    German translation: Es ist eine hintere Gelegenheit, dass es passiert.
    MQM annotations: 
    {"annotations": [{"error_span": "hintere", "category": "accuracy/mistranslation", "severity": 
    "minor", "is_source_error": "no"}, {"error_span": "es", "category": "accuracy/mistranslation", 
    "severity": "minor", "is_source_error": "no"}]}
        \end{Verbatim}
    \end{CJK}   
    
    \begin{CJK}{UTF8}{gbsn}
        \begin{Verbatim}
    English source: I love playing football with my friends.
    German translation: Der Hund läuft im Park.
    MQM annotations: 
    {"annotations": [{"error_span": "all", "category": "non-translation", "severity": "major", 
    "is_source_error": "no"}]}
        \end{Verbatim}
    \end{CJK} 
    }
\caption{Our examples used in Stage 1 for the Accuracy agent in En-De few-shot prompting.}
\label{app:fluen_examples}
\end{figure*}

\begin{figure*}[htb]
{\footnotesize
\begin{CJK}{UTF8}{gbsn}
    \begin{CJK}{UTF8}{gbsn}
        \begin{Verbatim}
    English source: I'll share with a couple of steps to perform into your device, okay?
    German translation: Ich werde mit ein paar Schritten teilen, um in Ihr Gerät zu spielen, okay?
    MQM annotations: 
    {"annotations": [{"error_span": "mit", "category": "fluency/grammar", "severity": "minor", 
    "is_source_error": "no"}]}
        \end{Verbatim}
    \end{CJK}

    \begin{CJK}{UTF8}{gbsn}
        \begin{Verbatim}
    English source: Thank you for contacting #PRS_ORG#, it was my pleasure to assist you today.
    German translation: Vielen Dank für die Kontaktaufnahme mit #PRS_ORG#, es war mir eine Freude, 
    Ihnen heute behilflich zu sein.
    MQM annotations: 
    {"annotations": []}
        \end{Verbatim}
    \end{CJK} 
        
    \begin{Verbatim}
    English source: I was scared that Covid was going to be worse.
    German translation: Ich hatte Angst, dass Covid schlimmer wird.
    MQM annotations: 
    {"annotations": [{"error_span": "wird", "category": "fluency/grammar", "severity": "minor", 
    "is_source_error": "no"}]}
        \end{Verbatim}
    \end{CJK}   
    
    \begin{CJK}{UTF8}{gbsn}
        \begin{Verbatim}
    English source: I love playing football with my friends.
    German translation: Der Hund läuft im Park.
    MQM annotations: 
    {"annotations": [{"error_span": "all", "category": "non-translation", "severity": "major", 
    "is_source_error": "no"}]}
        \end{Verbatim}
    \end{CJK} 
    }
\caption{Our examples used in Stage 1 for the Fluency agent in En-De few-shot prompting.}
\label{app:fluen_examples}
\end{figure*}

\begin{figure*}[htb]
{\footnotesize
\begin{CJK}{UTF8}{gbsn}
    \begin{CJK}{UTF8}{gbsn}
        \begin{Verbatim}
    English source: Deluxe Manual/ Battery Powered Vacuum Erection Penis Pump, manufactured by VVI 
    Ltd England, allows you to get a handle on your erectile dysfunction, commonly known as ED.
    German translation: Mit der von VVI Ltd England hergestellten Deluxe Manual / Battery Powered 
    Vacuum Erection Penis Pump können Sie Ihre erektile Dysfunktion, allgemein bekannt als ED, in 
    den Griff bekommen.
    MQM annotations: 
    {"annotations": []}
        \end{Verbatim}
    \end{CJK}

    \begin{CJK}{UTF8}{gbsn}
        \begin{Verbatim}
    English source: So far, 58,595,066 people have received the first dose of the COVID vaccine, 
    49,157,835 have received the second dosage and 2,237,841 have gotten the booster shots.
    German translation: Bisher haben 58.595.066 Menschen die erste Dosis des COVID-Impfstoffs erhalten, 
    49.157.835 haben die zweite Dosis erhalten und 2.237.841 haben die Auffrischimpfungen erhalten.
    MQM annotations: 
    {"annotations": []}
        \end{Verbatim}
    \end{CJK} 
        
    \begin{Verbatim}
    English source: I hope you have an excellent day.
    German translation: Ich wünsche Ihnen einen schönen Tag.
    MQM annotations: 
    {"annotations": [{"error_span": "wird", "category": "fluency/grammar", "severity": "minor", 
    "is_source_error": "no"}]}
        \end{Verbatim}
    \end{CJK}   
    
    \begin{CJK}{UTF8}{gbsn}
        \begin{Verbatim}
    English source: I love playing football with my friends.
    German translation: Der Hund läuft im Park.
    MQM annotations: 
    {"annotations": [{"error_span": "all", "category": "non-translation", "severity": "major", 
    "is_source_error": "no"}]}
        \end{Verbatim}
    \end{CJK} 
    }
\caption{Our examples used in Stage 1 for the Terminology agent in En-De few-shot prompting.}
\label{app:fluen_examples}
\end{figure*}

\begin{figure*}[htb]
{\footnotesize
\begin{CJK}{UTF8}{gbsn}
    \begin{CJK}{UTF8}{gbsn}
        \begin{Verbatim}
    English source: Iran reports lowest number of daily COVID-19 cases in more than one year
    German translation: Der Iran meldet die niedrigste Anzahl täglicher COVID-19-Fälle seit mehr
    als einem Jahr
    MQM annotations: 
    {"annotations": [{"error_span": "Anzahl", "category": "style/awkward", "severity": "minor", 
    "is_source_error": "no"}]}
        \end{Verbatim}
    \end{CJK}

    \begin{CJK}{UTF8}{gbsn}
        \begin{Verbatim}
    English source: Over the past 24 hours, 19 provinces reported almost no death case or only one dead.
    German translation: In den letzten 24 Stunden meldeten 19 Provinzen fast keinen Todesfall oder nur
    einen Toten.
    MQM annotations: 
    {"annotations": []}
        \end{Verbatim}
    \end{CJK} 
        
    \begin{Verbatim}
    English source: Shiba Inu is the latest meme-crypto to go viral and despite being down almost 60% 
    from it's all-time high, the market cap still stands at an eye-watering $20 billion, making it the 
    12th biggest crypto in the world by valuation.
    German translation: Shiba Inu ist die neueste Meme-Krypto, die viral wird, und obwohl sie fast 60% 
    von ihrem Allzeithoch entfernt ist, liegt die Marktkapitalisierung immer noch bei atemberaubenden 20 
    Milliarden US-Dollar und ist damit die 12. größte Krypto der Welt nach Bewertung.
    MQM annotations: 
    {"annotations": [{"error_span": "wird", "category": "style/awkward", "severity": "minor", 
    "is_source_error": "no"}]}
        \end{Verbatim}
    \end{CJK}   
    
    \begin{CJK}{UTF8}{gbsn}
        \begin{Verbatim}
    English source: I love playing football with my friends.
    German translation: Der Hund läuft im Park.
    MQM annotations: 
    {"annotations": [{"error_span": "all", "category": "non-translation", "severity": "major", 
    "is_source_error": "no"}]}
        \end{Verbatim}
    \end{CJK} 
    }
\caption{Our examples used in Stage 1 for the Style agent in En-De few-shot prompting.}
\label{app:fluen_examples}
\end{figure*}

\end{document}

%% file: table/intro.tex
\begin{table}[!htb]
\centering
\resizebox{\columnwidth}{!}{%
\begin{tabular}{lccc}
\toprule
Metric                    & Meta & System-Level & Segment-Level \\
\midrule
\multicolumn{4}{l}{\textcolor{gray}{\textit{LLM-as-a-judge:}}} \\
M-MAD (Ours)  & 0.808 (1)   & 0.987 & 0.517 \\
GEMBA-MQM & 0.784 (5)  & 0.986 & 0.472\\
GEMBA-DA  & 0.765 (8) & 0.990 & 0.474\\
EAPrompt  & 0.760 (9)  &  0.907 & 0.452\\
\cdashlinelr{1-4}
\multicolumn{4}{l}{\textcolor{gray}
{\textit{Automatic metrics:}}} \\
XCOMET-QE-Ensemble & 0.797 (2) & 0.892 & 0.533\\
\grayrow{}  MetricX-23  & 0.794 (3) & 0.873 & 0.531 \\
MetricX-23-QE & 0.787 (4) & 0.859 & 0.527\\
COMETKiwi & 0.776 (6)  & 0.963 & 0.525\\
\grayrow{}  MaTESe        & 0.773 (7) & 0.889 & 0.479 \\
\grayrow{}  COMET       & 0.754 (10)  &  0.898 & 0.514\\
\grayrow{}  BLEURT-20   & 0.751 (11)  &  0.880 & 0.518\\
\grayrow{}  BERTscore   & 0.720 (12)  &  0.810 & 0.499\\
\grayrow{}  chrF       & 0.671 (13)  &  0.809 & 0.485\\
\grayrow{}  BLEU       & 0.670 (14)  &  0.734 & 0.472\\
\toprule
\end{tabular}
}
\caption{Results of the WMT 2023 Metrics Shared Task ZH-EN testset. The first column displays the overall meta-evaluation performance. The second column shows the system-level Pearson correlation. The third column shows the score of segment-level Accuracy-t. Metrics with gray background is reference-based.}
\label{tab:intro}
\vspace{-5mm}
\end{table}

%% file: table/main_results.tex
\begin{table*}[t]
    \centering
    \setlength{\tabcolsep}{3pt}
    \small
    \begin{tabular}{p{3.5cm}p{1.1cm}p{1.1cm}p{1.1cm}p{1.1cm}p{1.1cm}c p{1.1cm}p{1.1cm}p{1.1cm}p{1.1cm}}
        \hline
        \toprule
        & & \multicolumn{4}{c}{\sc zh-en} & & \multicolumn{4}{c}{\sc en-de} \\ 
        \cmidrule{3-6} \cmidrule{8-11}
        \multirow{2}{*}{\sc Metric} & \multirow{2}{*}{\sc meta} & \multicolumn{2}{c}{System-Level} & \multicolumn{2}{c}{Segment-Level} & & \multicolumn{2}{c}{System-Level} & \multicolumn{2}{c}{Segment-Level} \\
        & & Acc & Pearson & Acc-t & Pearson & & Acc & Pearson & Acc-t & Pearson \\
        \midrule


        \multicolumn{11}{c}{\textit{Reference-based Metrics}}\\
        chrF & 0.695 & 0.762 & 0.809 & 0.485 & 0.063 & & 0.818 & 0.866 & 0.519 & 0.232 \\

        BLEU & 0.703 & 0.781 & 0.734 & 0.472 & 0.119 & & 0.894 & 0.917 & 0.520 & 0.192 \\
        BERTScore & 0.736 & 0.857 & 0.810 & 0.499 & 0.236 & & 0.879 & 0.891 & 0.528 & 0.325 \\
        MaTESe & 0.778 & 0.914 & 0.889 & 0.479 & 0.511 & & 0.848 & 0.918 & 0.528 & 0.554 \\
        COMET & 0.781 & 0.857 & 0.898 & 0.514 & 0.396 & & 0.970 & 0.990 & 0.574 & 0.432 \\
        BLEURT-20 & 0.782 & 0.857 & 0.880 & 0.518 & 0.378 & & 0.970 & 0.990 & 0.572 & 0.484 \\
        MetricX-23 & 0.808 & 0.895 & 0.873 & 0.531 & 0.625 & & 0.909 & 0.977 & 0.603 & 0.585 \\
        XCOMET-Ensemble & 0.826 & 0.905 & 0.927 & 0.543 & 0.650 & & 0.939 & 0.980 & 0.604 & 0.675 \\
       \hline
        \multicolumn{11}{c}{\textit{Reference-free Metrics}}\\
        \multicolumn{10}{l}{\textcolor{gray}{\textit{Learned Automatic Metrics}}}\\
        COMETKiwi & 0.793 & 0.876 & 0.963 & 0.525 & 0.442 & & 0.985 & 0.946 & 0.569 & 0.475 \\
        MetricX-23-QE & 0.806 & 0.867 & 0.859 & 0.527 & 0.647 & & 0.924 & 0.969 & 0.596 & 0.626 \\
        
        XCOMET-QE-Ensemble & 0.818 & 0.886 & 0.892 & 0.533 & 0.647& & 0.955 & 0.974 & 0.588 & 0.679 \\
        \cdashlinelr{1-11}
        \multicolumn{10}{l}{\textcolor{gray}{\textit{LLM-as-a-judge}}}\\
        
        EAPrompt & 0.772 & 0.876 & 0.907 & 0.452 & 0.516 & & 0.939 & 0.962 & 0.471 & 0.520 \\

        GEMBA-MQM & 0.784 & 0.933 & 0.986 & 0.472 & 0.475 & & 0.970 & 0.973 & 0.474 & 0.429 \\

        \textbf{M-MAD (Ours)} & 0.814 & 0.933 & 0.987 & 0.517 & 0.577 & & 0.970 & 0.979 & 0.555 & 0.552 \\

        \bottomrule 
    \end{tabular}
    \caption{
Results for system-level and segment-level WMT 23 Metrics Shared Task on ZH-EN and EN-DE. \textit{Ensemble} represents the best performance from XCOMET-XL and XCOMET-XXL. \textbf{M-MAD} outperforms all LLM-as-a-judge methods and achieves performance comparable to SoTA model-based automatic metrics.
    }
    \label{tab:main_results}
\end{table*}

%% file: table/ablation.tex
\begin{table}[ht]
\centering

\resizebox{\columnwidth}{!}{%
\begin{tabular}{lcccccccccc}
\toprule[0.5mm]
 \textbf{Ablations}  & \textbf{Meta} & \textbf{System-Level} & \textbf{Segment-Level} \\\midrule
 \raggedleft{w/o Stage 1 (DP)} & {-0.041} & {-0.038}  & {-0.145} \\
 \raggedleft{w/o Stage 2 (MAD)} & {-0.006} & {-0.019}  & {-0.002} \\
 \raggedleft{w/o Stage 3 (FJ)} & {-0.011} & {-0.038}  & {-0.021} \\
\bottomrule[0.5mm]
\end{tabular}
}
\caption{Ablation studies of three stages in M-MAD on WMT 23 Metrics Shared Task ZH-EN. The first column displays the overall meta-evaluation performance. The second column shows the system-level Accuracy. The third column shows the segment-level Pearson. }
\label{tab:ablation}
\end{table}

%% file: table/error_span.tex
\begin{table}[t]
\centering
\small
\resizebox{0.9\columnwidth}{!}{%
\begin{tabular}{lcccc}
\toprule[0.5mm]
\textbf{Method} & \textbf{Precision} & \textbf{Recall} & \textbf{F1 Score}\\  \midrule
\multicolumn{1}{l}{EAPrompt} & 0.29 & 0.38 & 0.33 \\
\multicolumn{1}{l}{GEMBA-MQM} & 0.28 & 0.54 & 0.37 \\
\multicolumn{1}{l}{Ours} & 0.41 & 0.78 & 0.54 \\
\bottomrule[0.5mm]
\end{tabular}
}
\caption{Error span prediction performance. Extracting error spans from unstructured LLM-as-a-judge methods like GEMBA-MQM and EAPrompt is highly challenging. We address this by matching their output strings to the gold annotations. This approach may lead to slightly inflated scores for these two methods.}
\label{tab:span_score} 
\end{table}

%% file: fig/distribution.tex
\definecolor{critical}{HTML}{a90674}
\definecolor{major}{HTML}{f6736b}
\definecolor{minor}{HTML}{fae1af}
\definecolor{noerror}{HTML}{d2e8f1}

\pgfplotsset{width=7cm, height=4cm,
    /pgfplots/ybar legend/.style={
    /pgfplots/legend image code/.code={%
      \draw[##1,/tikz/.cd,yshift=-0.25em]
        (0cm,0cm) rectangle (7pt,0.8em);},
  },}
    \flushleft
    \small
    \begin{tikzpicture}  
    \begin{groupplot}[
          group style={
          group name=plot,
          horizontal sep=0pt,
          vertical sep=0pt,
          group size=3 by 1},]
      \nextgroupplot[
            xbar stacked,
            xmin=-0.5, xmax=20,
            bar width=10pt,
            ytick={0, 1, 2, 3},
            yticklabels={
            \texttt{GEMBA-MQM},
            \texttt{EAPrompt},
            \texttt{Ours},
            \texttt{Gold},
            },
            xtick={0, 4, 8, 12, 16, 20},
            xticklabels={\texttt{0}, \texttt{20}, \texttt{40}, \texttt{60}, \texttt{80}, \texttt{100}},
            ytick style={draw=none},
            ylabel style={align=left,
                        width=0pt,
                        },
            axis line style={draw=none},
            legend cell align=left,
            legend style={
                /tikz/column 2/.style={column sep=6pt},
        at={(0.5,0.1)},  
        anchor=south,  
                column sep=1.1ex,
                font=\small,
                draw=none,
                yshift=2.25cm,
            },
            legend columns=-1,
            ]

        \addplot[xbar, fill=minor,  postaction={}] coordinates {
            (3.746813933729822, 3) %
            (4.77909940526763, 2) %
            (2.403355989804588, 1) %
            (4.348980458793543, 0) %
        };
        \addplot[xbar, fill=major!80,  postaction={}] coordinates {
            (10.749787595581989, 3) %
            (10.807136788445199, 2) %
            (3.4643160577740018, 1) %
            (2.6635514018691584, 0) %
        };
        \addplot[xbar, fill=critical!60,  postaction={}] coordinates {
            (5.5033984706881895, 3) %
            (4.413763806287171, 2) %
            (14.132327952421411, 1) %
            (12.987468139337299, 0) %
        };

        \legend{
            \texttt{HQ},
            \texttt{MQ},
            \texttt{LQ}
            }
        ]
        ]
    \end{groupplot}
    \node[below=0.45cm of plot c1r1.south, anchor=north] {\small{Percentage (\%)}}; 
    \end{tikzpicture} 

%% file: table/debate_outside.tex
\begin{table}[ht]
\centering

\resizebox{\columnwidth}{!}{%
\begin{tabular}{lccc}
\toprule[0.5mm]

\textbf{Methods} & \textbf{Decoupled Multidimension} & \textbf{Debating} & \textbf{Meta} \\
\midrule
 GEMBA-MQM & \xmark & \xmark  & {0.784}  \\
\citet{liang-etal-2024-encouraging} & \xmark & \cmark & {0.665}  \\
 \citet{khandebating}$^{\heartsuit}$ & \xmark & \cmark  & {0.728}\\
 \citet{khandebating}$^{\spadesuit}$ & \xmark & \cmark & {0.712}  \\

M-MAD (Ours) & \cmark & \cmark  & {0.808}  \\

\bottomrule[0.5mm]
\end{tabular}
}
\caption{Applying the off-the-shelf debating framework directly to MT evaluation. The debating process is initialized from GEMBA-MQM ZH-EN, focusing on coupled MQM evaluation results. {$\heartsuit$} and {$\spadesuit$} denote two different strategies in \citet{khandebating}.}
\vspace{-3mm}
\label{tab:debating_frame}
\end{table}

%% file: table/cat_coarse.tex
\begin{table}[t]
\centering
\resizebox{\columnwidth}{!}{%
\begin{tabular}{lcccccc}
\toprule[0.5mm]
& & \multicolumn{2}{c}{\textbf{System-Level}} & \multicolumn{2}{c}{\textbf{Segment-Level}} \\
\cmidrule(lr){3-4} \cmidrule(lr){5-6}
\textbf{Topics} & \textbf{Meta} & \textbf{Acc} & \textbf{Pearson} & \textbf{Acc-t} & \textbf{Pearson} \\\midrule
Severity & {0.808} & {0.933} & {0.987} & {0.517} & {0.577} \\
Category & {0.666} & {0.819} & {0.444} & {0.423} & {0.320} \\
Entirety & {0.737} & {0.857} & {0.894} & {0.423} & {0.439} \\
\bottomrule[0.5mm]
\end{tabular}%
}
\caption{Comparison with debating on different topics on ZH-EN. Severity: debates on error severity. Category: debates on category correctness (e.g., fluency, accuracy). Entirety: debates without specific focus.}
\label{tab:debate_topic}
\end{table}

%% file: table/debate_strategies.tex
\begin{table}[ht]
\centering
\resizebox{\columnwidth}{!}{%
\begin{tabular}{lcccccc}
\toprule[0.5mm]
& & \multicolumn{2}{c}{\textbf{System-Level}} & \multicolumn{2}{c}{\textbf{Segment-Level}} \\
\cmidrule(lr){3-4} \cmidrule(lr){5-6}
\textbf{Methods} & \textbf{Meta} & \textbf{Acc} & \textbf{Pearson} & \textbf{Acc-t} & \textbf{Pearson} \\\midrule
Baseline & {0.802} & {0.914} & {0.975} & {0.519} & {0.575} \\
\cdashlinelr{1-6}
Consensus  & {0.808} & {0.933} & {0.987} & {0.517} & {0.577} \\
Deliberation & {0.805} & {0.924} & {0.978} & {0.520} & {0.574} \\
Interactive Review & {0.798} & {0.905} & {0.978} & {0.518} & {0.561} \\
Consultancy Review & {0.790} & {0.886} & {0.973} & {0.513} & {0.551} \\
\bottomrule[0.5mm]
\end{tabular}%
}
\caption{Comparison with different debating strategies on ZH-EN. Baseline is the same as Table~\ref{tab:ablation} w/o Stage 2.}
\label{tab:debate_strategy}
\end{table}

%% file: table/good_case.tex
\begin{table*}[t]
    \centering
    \small
    \begin{tabular}{lp{13cm}}
    \toprule
    \multirow{2}{*}{Source} & \chinese{表的走时是很准的，误差完全在可接受范围内，表的包装里有珠宝鉴定证书，表盘的12字那里是真钻石，这个自己戴很合适，作为礼品送出去也很有面子。} \\
    \cdashlinelr{1-2}
    \multirow{3}{*}{Reference} & The watch's trip time is extremely accurate, and its error is entirely within acceptable bounds, the 12 on the dial is a genuine diamond, and a certificate of jewelry appraisal is included with the watch, both as a present and as something you may wear by yourself, this is excellent.\\
    \cdashlinelr{1-2}
    \multirow{3}{*}{Translation} & The time of the watch is very accurate, and the error is completely within the acceptable range. There is a jewelry appraisal certificate in the packaging of the watch, and there is a real diamond on the 12 number on the dial. \\
    \cdashlinelr{1-2}
    \multirow{1}{*}{Gold Annotations} &\hlred{Minor}: \hlblue{style/unnatural or awkward} - "time", "on", "12", "number". (Score: -4)\\
    \hline
    \multirow{3}{*}{EAPrompting} &\hlred{Major}:
\hlblue{accuracy/mistranslation} - "the 12 number"
\hlblue{terminology/inappropriate for context} - "on the dial" \hlred{Minor}:
\hlblue{style/unnatural or awkward} - "the time of the watch", "there is a real diamond", "on the 12 number".  (Score: -13)\\
    \cdashlinelr{1-2}
    \multirow{3}{*}{GEMBA-MQM} &\hlred{Major}:
\hlblue{accuracy/mistranslation} - "12 number on the dial"
\hlblue{style/awkward} - "the error is completely within the acceptable range". (Score: -10) \\
    \cdashlinelr{1-2}
    \multirow{3}{*}{M-MAD Stage 1} & 
\hlred{Major:}
\hlblue{accuracy/mistranslation} - "there is a real diamond on the 12 number on the dial"
\hlblue{accuracy/omission translation} - "this is suitable for wearing", "as a gift, it is also very impressive"
\hlred{Minor:}
\hlblue{fluency/grammar} - "the 12 number"
\hlblue{terminology/inappropriate for context} - "12 number on the dial"
\hlblue{style/awkward} - "the 12 number on the dial".\\
    \cdashlinelr{1-2}
    \multirow{3}{*}{M-MAD Stage 2} & 
\hlred{Minor}:
\hlblue{accuracy/mistranslation} - "there is a real diamond on the 12 number on the dial"
\hlblue{accuracy/omission} - "this is suitable for wearing", "as a gift, it is also very impressive"
\hlblue{fluency/grammar} - "the 12 number"
\hlblue{terminology/inappropriate for context} - "12 number on the dial"
\hlblue{style/awkward} - "the 12 number on the dial".\\
    \cdashlinelr{1-2}
    \multirow{3}{*}{M-MAD Stage 3} & 
\hlred{Minor}:
\hlblue{accuracy/mistranslation} - "there is a real diamond on the 12 number on the dial"
\hlblue{accuracy/omission} - "this is suitable for wearing", "as a gift, it is also very impressive"
\hlblue{style/awkward} - "the 12 number on the dial". (Score: -4)
    \\
    \midrule

    \end{tabular}
    \caption{\label{tab:good_case_study} Case study. \hlred{Red} denotes severity, and \hlblue{blue} denotes the category and subcategory.}
    \end{table*}

%% file: table/appendix_he.tex
\begin{table}[h]
    \centering
    \setlength{\tabcolsep}{3pt}
    \resizebox{\columnwidth}{!}{%
    \begin{tabular}{p{3.5cm}p{1.1cm}p{1.1cm}p{1.1cm}p{1.1cm}p{1.1cm}}
        \toprule
        & \multicolumn{5}{c}{\sc he-en} \\ 
        \cmidrule{2-6}
    \multirow{2}{*}{\sc Metric} & \multirow{2}{*}{\sc meta} & \multicolumn{2}{c}{System-Level} & \multicolumn{2}{c}{Segment-Level} \\
        & & Acc & Pearson & Acc-t & Pearson \\
        \midrule
        chrF & 0.695 & 0.821 & 0.776 & 0.460 & 0.221 \\
        BLEU & 0.683 & 0.795 & 0.769 & 0.442 & 0.220 \\
        BERTScore & 0.754 & 0.885 & 0.895 & 0.515 & 0.335 \\
        MaTESe & 0.792 & 0.936 & 0.906 & 0.550 & 0.459 \\
        COMET & 0.775 & 0.897 & 0.940 & 0.532 & 0.401 \\
        BLEURT-20 & 0.763 & 0.872 & 0.937 & 0.519 & 0.382 \\
        MetricX-23 & 0.807 & 0.923 & 0.910 & 0.577 & 0.548 \\
        COMETKiwi & 0.760 & 0.872 & 0.860 & 0.544 & 0.387 \\
        MetricX-23-QE & 0.788 & 0.897 & 0.858 & 0.564 & 0.520 \\
        XCOMET-QE-Ensemble & 0.789 & 0.897 & 0.909 & 0.554 & 0.498 \\
        \textbf{M-MAD (Ours)} & 0.787 & 0.910 & 0.972 & 0.538 & 0.425 \\
        \bottomrule 
    \end{tabular}
    }
    \caption{Results for system-level and segment-level WMT 23 Metrics Shared Task on HE-EN.}
    \label{tab:he_results}
\end{table}

%% file: table/bad_case.tex

\begin{table*}[h]
    \centering
    \small
    \begin{tabular}{lp{12cm}}
    \toprule
    \multirow{2}{*}{Source} & \chinese{你永远主动联系不上这个专员，也不知道她的工号，也没有直线联系电话，就是你联系不上她，只有她联系你。} \\
    \cdashlinelr{1-2}
    \multirow{3}{*}{Reference} & Since you don't know the commissioner's job number and there isn't a direct phone number to call, you'll never make the effort to get in touch with her, She is the only one who can reach you, You can't. \\
    \cdashlinelr{1-2}
    \multirow{3}{*}{Translation} & You never get in touch with the Commissioner, you don't know her work number, you don't have a direct line, you don't get in touch with her, she gets in touch with you. \\
    \cdashlinelr{1-2}
    \multirow{3}{*}{Gold Annotations} &{\hlred{Major}: \hlblue{accuracy/mistranslation} - "You", "never get", "you", "her", "you", "direct line", "you", "her", "she", "you". \hlred{Minor}: \hlblue{fluency/grammar} - "don", "don", "don", "with her, she", "gets"; \hlblue{fluency/spelling} - "Commissioner"; \hlblue{style/awkward} - "you don't get in touch with her, she gets in touch with you"; \hlblue{terminology/inappropriate} - "Commissioner".} (Score: -56)\\
    \cdashlinelr{1-2}
    \multirow{3}{*}{M-MAD Evaluation} & 
    {
        \hlred{Minor}: \hlblue{style/awkward} - "you don't get in touch with her, she gets in touch with you";
        \hlblue{terminology/inappropriate} - "Commissioner". (Score: -2)
    }
    \\
    \midrule

    \end{tabular}
    \caption{\label{tab:bad_case_study} Bad case study. Some gold annotations appear to be incorrectly labeled, making evaluation challenging. }
    \end{table*}

%% file: table/dataset.tex
\begin{table*}[ht]
\centering
\resizebox{\textwidth}{!}{%
\begin{tabular}{lccccl}
\toprule
\textbf{Dataset} &
  \textbf{Language Pair} &
  \textbf{Segments} &
  \textbf{Systems} &
  \textbf{Total Segments} &
  \textbf{Systems} \\ 
\midrule
\multirow{3}{*}{WMT23} &
  En-De &
  557 &
  14 &
  7798 &
  \begin{tabular}[c]{l}AIRC, GPT4-5shot, Lan-BridgeMT, NLLB\_Greedy, NLLB\_MBR\_BLEU, ONLINE-A, ONLINE-B,\\ ONLINE-G, ONLINE-M, ONLINE-W, ONLINE-Y, ZengHuiMT, refA, synthetic\_{ref}\end{tabular} \\ 
&
  He-En &
  1910 &
  14 &
  26740 &
  \begin{tabular}[c]{l}GTCOM\_Peter, GPT4-5shot, Lan-BridgeMT, NLLB\_Greedy, NLLB\_MBR\_BLEU, ONLINE-A, ONLINE-B,\\ ONLINE-G, ONLINE-Y, ZengHuiMT, Samsung\_Research\_Philippines, UvA-LTL, refA, refB\end{tabular} \\ 
&
  Zh-En &
  1976 &
  17 &
  33592 &
  \begin{tabular}[c]{l}ANVITA, GPT4-5shot, Lan-BridgeMT, NLLB\_Greedy, NLLB\_MBR\_BLEU, ONLINE-A, ONLINE-B,\\ ONLINE-G, ONLINE-M, ONLINE-W, ONLINE-Y, ZengHuiMT, HW-TSC, IOL\_Research, Yishu, refA, synthetic\_{ref}\end{tabular} \\ 
\bottomrule
\end{tabular}%
}
\caption{Statistics of testset. Source, reference texts, and translations are from the WMT23 Metrics Shared Task.}
\label{tab:mqm_stat}
\end{table*}